\documentclass{article}

\usepackage{microtype}
\usepackage{graphicx}
\usepackage{subfigure}
\usepackage{booktabs} 

\include{defns}
\usepackage{hyperref}
\usepackage[accepted]{icml2020}
\usepackage{amsmath}
\usepackage{amsmath, latexsym}
\usepackage{amsfonts}
\usepackage{mathtools}
\usepackage{amsthm}
\usepackage{dsfont}
\usepackage{booktabs}
\usepackage{xfrac}
\usepackage{bbm}
\usepackage{xparse}
\usepackage{lipsum}
\usepackage{enumerate}

\newcommand{\citet}{\cite}
\newcommand{\citep}{\cite}

\newcommand{\logit}{\mbox{logit}}
\def\paradot#1{\paragraph{#1.}}

\newcommand{\geo}{\text{\sc geo}}

\newcommand{\norm}[1]{\left\Vert #1 \right\Vert}

\newcommand{\gln}{\textsc{GLN}}

\newcommand{\cB}{\mathcal B}
\newcommand{\cC}{\mathcal C}

\newcommand{\cS}{\mathcal S}

\newcommand{\cW}{\mathcal W}

\newcommand{\cZ}{\mathcal Z}

\newcommand{\R}{\mathbb R}

\def\eps{\varepsilon}           

\allowdisplaybreaks 

\begin{document}

\twocolumn[

    \icmltitle{Gated Linear Networks}
    \icmlsetsymbol{equal}{*}
    
    \begin{icmlauthorlist}
    \icmlauthor{Joel Veness}{equal,DeepMind}
    \icmlauthor{Tor Lattimore}{equal,DeepMind}
    \icmlauthor{David Budden}{equal,DeepMind}
    \icmlauthor{Avishkar Bhoopchand}{equal,DeepMind}
    \icmlauthor{Christopher Mattern}{DeepMind}
    \icmlauthor{Agnieszka Grabska-Barwinska}{DeepMind}
    \icmlauthor{Eren Sezener}{DeepMind}
    \icmlauthor{Jianan Wang}{DeepMind}
    \icmlauthor{Peter Toth}{DeepMind}
    \icmlauthor{Simon Schmitt}{DeepMind}
    \icmlauthor{Marcus Hutter}{DeepMind}
    \end{icmlauthorlist}
    
    \icmlaffiliation{DeepMind}{DeepMind}
    \icmlkeywords{Gated Linear Networks}
    
    \vskip 0.3in

]
\printAffiliationsAndNotice{\icmlEqualContribution}

\begin{abstract}
This paper presents a new family of backpropagation-free neural architectures, Gated Linear Networks (GLNs). What distinguishes GLNs from contemporary neural networks is the distributed and local nature of their credit assignment mechanism; each neuron directly predicts the target, forgoing the ability to learn feature representations in favor of rapid online learning. Individual neurons can model nonlinear functions via the use of data-dependent gating in conjunction with online convex optimization. We show that this architecture gives rise to universal learning capabilities in the limit, with effective model capacity increasing as a function of network size in a manner comparable with deep ReLU networks. Furthermore, we demonstrate that the GLN learning mechanism possesses extraordinary resilience to catastrophic forgetting, performing comparably to a MLP with dropout and Elastic Weight Consolidation on standard benchmarks. These desirable theoretical and empirical properties position GLNs as a complementary technique to contemporary offline deep learning methods. 
\end{abstract}

\section{Introduction}

Backpropagation has long been the de-facto credit assignment technique underlying the successful training of popular neural network architectures such as convolutional neural networks and multilayer perceptions (MLPs). 
It is well known that backpropagation enables these networks to learn highly-relevant task-specific features. 
However, this method is not without its limitations. 
Contemporary neural networks trained via backpropagation require many epochs of training over massive datasets, limiting their effectiveness for data-efficient online learning. 
Interpretibility limitations can also prevent their application in domains where a human understandable solution is a mandatory requirement.
Their effectiveness is further limited in the continual learning setting by their tendency to catastrophically forget previously learnt tasks. 
Although various meta-learning \cite{peedrrrrooo} algorithms such as Elastic Weight Consolidation \citep[EWC]{Kirkpatrick17} have been effective in compensating for these limitations, it is interesting to explore whether alternative methods of credit assignment can give rise to complementary neural models with different strengths and weaknesses.

This paper introduces one alternative model family, Gated Linear Networks (GLNs), and studies their contrasting properties. 
The distinguishing feature of a GLN is its distributed and local credit assignment mechanism. 
This technique is a generalization of the PAQ family \cite{Mahoney2000, Mahoney2005, Mahoney2013} of online neural network models, which are well-known in the data compression community for their excellent sample efficiency \cite{Mahoney2013, cmix}. By interpreting these systems within an online convex programming \cite{zinkevich03} framework as a sequence of data dependent linear networks coupled with a choice of gating function, we are able to provide a new algorithm and gating mechanism that opens up their usage to the wider machine learning community.

GLNs have a number of desirable properties.
Their local credit assignment mechanism is derived by associating a separate convex loss function to each neuron, which greatly simplifies parameter initialization and optimization, and provides significant sample efficiency benefits when learning online.
Importantly, we show that these benefits do not come at the expense of capacity in practice, which adds further weight to previously obtained asymptotic universality results \cite{Veness:17}.

GLNs possess excellent online learning capabilities, which we demonstrate by showing performance competitive with batch-trained MLPs on a variety of standard classification, regression and density modeling tasks, using only a single online pass through the data.
In terms of interpretibility, we show how the data-dependent linearity of the predictions can be exploited to trivialise the process of constructing meaningful saliency maps, which can be of great reassurance to practitioners that the model is predicting well for the right reasons.
Perhaps most interestingly, we demonstrate that our credit assignment mechanism is extraordinarily resilient to catastrophic forgetting, achieving performance competitive with EWC on a standard continual learning benchmark with no knowledge of the task boundaries.

\section{Background}

\begin{figure*}[t!]
	\centering
	\vspace{-2em}
	\includegraphics[scale=0.45]{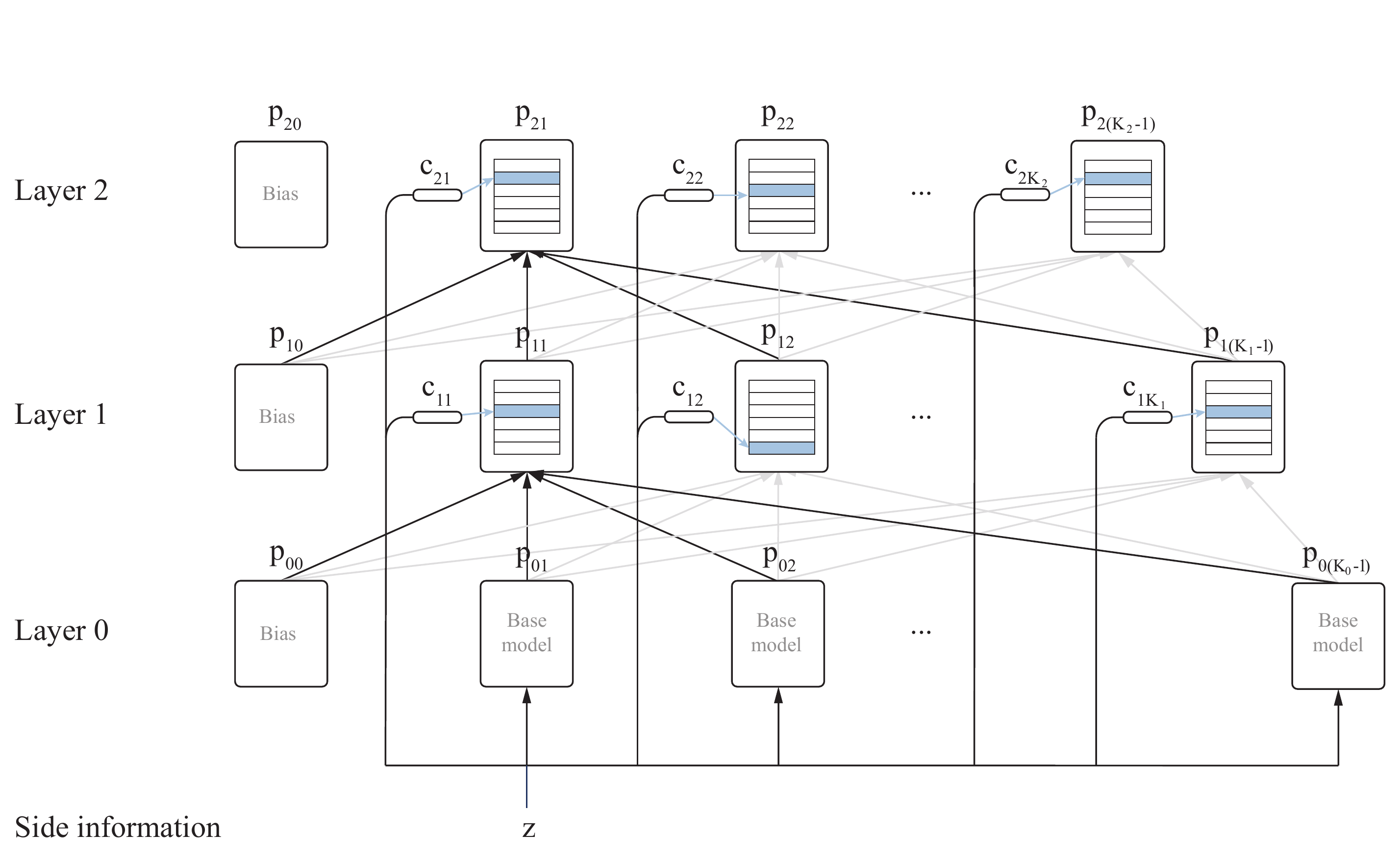}
	\vspace{-0.5em}
    \caption{\centering A graphical depiction of a Gated Linear Network. Each neuron receives inputs from the previous layer as well as the broadcasted side information $z$. The side information is passed through all the gating functions, whose outputs $s_ij=c_{ij}(z)$ determine the active weight vectors (shown in blue).}
    \label{fig:gln}
\end{figure*}

In this section we review some necessary background on geometric mixing, a parametrised way of combining probabilitstic forecasts, and show how to adapt its parameters using online convex programming.
Later we will combine this method with a gating mechanism to define a single neuron within a GLN.

\paragraph{Geometric Mixing.}

Geometric Mixing is a simple and well studied ensemble technique for combining probabilistic forecasts.
It has seen extensive application in statistical data compression \cite{Mattern12,Mattern13}.
Given $p_1, p_2, \dots, p_d$ input probabilities predicting the occurrence of a single binary event, geometric mixing predicts $\sigma(w^\top \sigma^{-1}(p))$, where $\sigma(x):=1/(1+{\rm e}^{-x})$ denotes the sigmoid function, $\sigma^{-1}$ defines the logit function, $p := (p_1, \dots, p_d)$ and $w \in \mathbb{R}^d$ is the weight vector which controls the relative importance of the input forecasts.
One can easily show the following identity:
\begin{equation*}
\sigma\left(w^\top \sigma^{-1}(p)\right) = \frac{\prod_{i=1}^d p_{i}^{w_i}}{ \prod_{i=1}^d p_{i}^{w_i} + \prod_{i=1}^d (1 - p_{i})^{w_i} }, 
\end{equation*}
which makes it clear that that geometric mixing implements a type of product of experts \cite{Hinton2002} operation.
This leads to the following interesting properties: 
setting $w_i = 1/d$ is equivalent to taking the geometric mean of the $d$ input probabilities;
if the $j$th component of $w_j$ is 0 then the contribution of $p_j$ is ignored, and if $w = 0$ then the geometric mixture predicts $1/2$;
and finally, due to the product formulation, every forecaster has ``the right of veto", in the sense that a single $p_{i}$ close to 0 coupled with a $w_i > 0$ drives the geometric mixture prediction close to zero.

\paragraph{Online Convex Programming Formulation.}
 
We now describe how to adapt the geometric mixing parameters using online convex programming \cite{zinkevich03,Hazan:16}.
Let $\cB := \{ 0,1 \}$.
As we are interested in probabilistic prediction, we assume a standard online learning framework for the logarithmic loss, where at each round $t \in \mathbb{N}$ a predictor outputs a binary distribution $q_t~:~\cB~\to~[0,1]$, with the environment responding with an observation $x_t \in \cB$, causing the predictor to suffer a loss $\ell_t(q_t, x_t) = -\log q_t(x_t)$ before moving onto round $t+1$.

In the case of geometric mixing, we first define our parameter space to be a non-empty convex set $\cW \subset \mathbb{R}^d$.
As the prediction depends on both the $d$ dimensional input predictions $p_t$ and the parameter vector $w \in \cW$, we abbreviate the loss at time $t$, given target $x_t$, using parameters $w$ by 
\begin{align}\label{eq:geo_loss_defn}
\ell^{\geo}_t(w) := -\log\left( \geo_{w}(x_t \,;\, p_t) \right), 
\end{align}
with $\geo_{w}( 1; p_t) := \sigma(w^\top \sigma^{-1}(p_t))$ and 
$\geo_{w}(0 \,;\, p_t) := 1 - \geo_{w}(1 \,;\, p_t)$.
One can show that $\ell^\geo_t(w)$ is a convex function of $w$ \cite{Mattern13} and that the gradient of the loss with respect to $w$ is given by
\begin{equation}\label{eq:grad_loss}
\nabla \ell^{\geo}_t(w) = \left( {\geo}_w( 1 ; p_t) - x_t \right) \logit(p_t).
\end{equation}
Furthermore we can bound the 2-norm of the gradient of the loss with 
$\norm{\nabla \ell^\geo_t(w)}_2 \leq \sqrt{d} \log \left( \frac{1}{\eps} \right)$ provided that $p_t \in [\eps, 1 - \eps]^{d}$ for some $\eps \in (0,1/2)$ for every time $t$.
These properties of the sequence of loss functions make it possible to apply one of the many different online convex programming techniques to adapt $w$ at the end of each round.
In this paper we restrict our attention to Online Gradient Descent \citep{zinkevich03}, with $\cW$ equal to some choice of hypercube, for reasons of computational efficiency.
This gives a $O(\sqrt{T})$ regret bound with respect to the best $w^* \in \cW$ chosen in hindsight provided an appropriate schedule of decaying learning rates is used.

\section{Gated Geometric Mixing}

We define the GLN neuron as a \emph{gated geometric mixer}, which we obtain by adding a contextual gating procedure to geometric mixing.
Here, contextual gating has the intuitive meaning of mapping particular input examples to particular sets of weights.
The key change compared with normal geometric mixing is that now our neuron will also take in an additional type of input, \emph{side information}, which will be used by the contextual gating procedure to determine an active subset of the neurons weights to use for a given example.
In typical applications the side information will simply be the input features associated with a given example.

More formally, associated with each neuron is a context function $c : \cZ \to \cC$, where $\cZ$ is the set of possible \textit{side information} and $\cC = \{0,\dots,k-1 \}$ for some $k \in \mathbb{N}$ is the \textit{context space}.
Given a convex set $\cW \subset \mathbb{R}^d$, each neuron is parametrized by a matrix
$
W = \begin{bmatrix}
w_0 \dots w_{k-1} 
\end{bmatrix}^\top
$
with each row vector $w_i \in \cW$ for $0 \leq i < k$.
The context function $c$ is responsible for mapping a given piece of side information $z_t \in \cZ$ to a particular row $w_{c(z_t)}$ of $W$, which we then use with standard geometric mixing.

In other words, a Gated Geometric Mixer can be defined in terms of geometric mixing as
\begin{equation}
\label{eq:gated_geo_defn}
\geo^c_{W}(x_t \,;\, p_t, z_t) := \geo_{w_{c(z_t)}}(x_t \,;\, p_t),
\end{equation}
with the associated loss function $-\log \left(\geo^c_{W}(x_t \,;\, p_t, z_t)\right)$ inheriting all the properties needed to apply Online Convex Programming directly from Equation \ref{eq:geo_loss_defn}.
The key intuition behind gating is that it allows each neuron to be able to specialize its weighting of input predictions based on some particular property of the side information.

\paragraph{Universal context functions.}
We now introduce a \emph{halfspace gating} mechanism that is tailored towards machine learning applications whose input features lie in $\mathbb{R}^d$. 
Although not the focus of this work, its worth noting that this choice gives rise to universal approximation capabilities for sufficiently large GLNs \cite{Veness:17}.
Once we are in a position to describe the learning dynamics of multiple interacting neurons, the rationale for this class of context functions will become more clear.
Exploring alternative gating mechanisms is an exciting area for future work.

\paragraph{Halfspace gating.} 
Given a normal $v \in \R^d$ and offset $b \in \R$, consider the associated affine hyperplane $\{x \in \mathbb{R}^d : x \cdot v = b \}$.
This divides $\mathbb{R}^d$ in two, giving rise to two half-spaces, one of which we denote $$H_{v,b} = \{x \in \mathbb{R}^d : x \cdot v \geq b\}.$$
The associated half-space context function is then given by $\mathds{1}_{H_{v,b}}(z)$, where $\mathds{1}_\cS(s) := 1$ if $s \in \cS$ and 0 otherwise.

\paragraph{Context composition.}
Richer notions of context can be created by composition.
In particular, any finite set of $m$ context functions $\{ c_i : \cZ \to \cC_i \}_{i=1}^m$ with associated context spaces $\cC_1, \dots, \cC_m$ can be composed into a single higher order context function $c : \cZ \to \cC$, 
where $\cC = \cC_1\times...\times\cC_d\cong\{0,...,|\cC|-1\}$ by defining $c(z) = (c_1(z),...,c_d(z))$.

For example, we could combine $m=4$ different halfspace context functions into a single context function with a context space containing $|\cC|=16$ elements.
From here onwards, whenever this technique is used, we will refer to the choice of $m$ as the \emph{context dimension}.

\section{Gated Linear Networks}

We now introduce \emph{Gated Linear Networks}, which are feed-forward networks composed of many layers of gated geometric mixing neurons as shown in Figure \ref{fig:gln}.
Each neuron in a given layer outputs a gated geometric mixture of the predictions from the previous layer, with the final layer consisting of just a single neuron.
In a supervised learning setting, a GLN is trained on (side information, base predictions, label) triplets $(z_t,p_t,x_t)_{t=1,2,3,...}$ derived from input-label pairs $(z_t, x_t)$.
There are two types of input to neurons in the network: the first is the side information $z_t$, which can be thought of as the input features; the second is the input to the neuron, which will be the predictions output by the previous layer, or in the case of layer 0, some (optionally) provided base predictions $p_t$ that typically will be a function of $z_t$. 
Each neuron will also take in a constant bias prediction, which helps empirically and is essential for universality guarantees \cite{Veness:17}.

\paragraph{GLN architecture.}
A GLN is a network of gated geometric mixers organized in $L+1$ layers indexed by $i \in \{0,\ldots,L\}$, with $K_i$ models in each layer.
Neurons are indexed by their position in the network when laid out on a grid; for example, neuron $(i,k)$ will refer to the $k$th neuron of the $i$ layer and $p_{ik}$ will refer to the output of neuron $(i,k)$.
The output of layer $i$ will be denoted by $p_i$.
The zeroth layer of the network is called the \textit{base layer}, whose output $p_0$ will typically be instantiated via scaling or squashing each component of the current side information $z$ to lie within $[\eps,1-\eps]$.
The nonzero layers are composed of gated geometric mixing neurons. 
Associated to each of these will be a fixed context function $c_{ik} : \cZ \to \cC$ that determines the behavior of the gating at neuron $(i,k)$. 
In addition to the context function, for each context $c \in \cC$ and each neuron $(i,k)$ there is an associated weight vector $w_{ikc} \in \R^{K_{i-1}}$ which is used to geometrically mix the inputs whenever active.
The bias outputs $p_{i0}$ for $0 \leq i \leq L$ can be set to be any constant $\beta \in [\eps, 1-\eps] \setminus \{ 0.5 \}$.
Given a $z \in \cZ$, a weight vector for each neuron is determined by evaluating its associated context function.
For layers $i \geq 1$, the $k$th node in the $i$th layer receives as input the vector $p_{i-1}$ of dimension $K_{i-1}$ of predictions of the preceding layer.

\paragraph{GLNs are data dependent linear networks.}
Without loss of generality, here we assume that the network is estimating the probability of the target being positive.
The output of a single neuron is the geometric mixture of the inputs with respect to a set of weights that depend on its context, namely
\begin{align*}
p_{ik}(z) = \sigma \left( w_{ikc_{ik}(z)} \cdot \sigma^{-1} \left( p_{i-1} \left( z \right) \right) \right). 
\end{align*}
The output of layer $i$ can be written in matrix form as
\begin{align}\label{eq:neuron}
p_i(z) = \sigma(W_i(z) \, \sigma^{-1}(p_{i-1}(z)))\,,
\end{align}
where $W_i(z) \in \R^{K_i \times K_{i-1}}$ is the matrix with $k$th row equal to $w_{ik}(z) = w_{ikc_{ik}(z)}$.
Iterating Equation \ref{eq:neuron} once gives 
\begin{equation*}
p_i(z) = \sigma\left( W_i \left(z \right) \sigma^{-1} \left( \sigma \left( W_{i-1} \sigma^{-1}\left(p_{i-2}\left(z \right) \right) \right) \right) \right).
\end{equation*}
Observing that the logit and sigmoid functions cancel,  simplifying the $i$th iteration of Equation \ref{eq:neuron} gives
\begin{align}
\label{eq:linear}
p_i(z)= \sigma\Bigl( W_{i}(z) W_{i-1}(z) \dots W_{1}(z) \, \sigma^{-1}(p_0(z)) \Bigr),
\end{align}
which shows the network behaves like a linear network \citep{Baldi1989,SaxeMG13}, but with weight matrices that are data-dependent.
Without the data dependent gating, the product of matrices would collapse to a single linear mapping and provide no additional modeling power over a single neuron \citep{minsky69perceptrons}.
 
\paragraph{Local learning in GLNs.}

We now describe how the weights are learnt in a Gated Linear Network using Online Gradient Descent (OGD) \citep{zinkevich03} locally at each neuron.
They key observation is that as each neuron $(i,k)$ in layers $i > 0$ is itself a gated geometric mixture, all of these neurons can be thought of as individually predicting the target.
Thus given side information $z$ and from Equations \ref{eq:geo_loss_defn} and \ref{eq:gated_geo_defn}, each neuron $(i,k)$ suffers a loss convex in its active weights $u := w_{ik{c_{ik}(z)}}$ of
$$
  \ell_t(u) :=~ -\log \left( \geo_{u} \left(x_t \,; \, p_{i-1}\right) \right).
$$
Algorithmically, a single step of OGD consists of two parts: a gradient step, and then a projection back into some convex weight space $\cW$. 
The gradient step can be trivially obtained from Equation \ref{eq:grad_loss}.
It is well known that the projection step can be implemented via clipping if the convex set $\cW$ is a scaled hypercube.
In our case this can be achieved if we force every component of each weight vector, for each neuron, to lie within $[-b,b]$ for some constant $b > 1$. 

\paragraph{Weight initialization.}
One benefit of a convex loss is that weight initialization is less important in determining overall model performance, and one can safely recommend deterministic initialization schemes that favor reproducibility of results.
While other choices are possible, we found the initialization $w_{ikc} = 1/K_{i-1}$ for all $i,k,c$ to be a good choice empirically, which causes geometric mixing to initially compute a geometric average of its input.

\paragraph{Algorithm.}

\begin{algorithm}[t!]
\caption{$\gln(\Theta, z, p, x, \eta, \text{update})$.\\
\vspace{0.2em}
Perform a forward pass and optionally update weights.
\vspace{0.2em}
}

\label{algo:updategln}
\begin{algorithmic}[1]
    \vspace{0.4em}
    \STATE {\bfseries Input:} GLN weights $\Theta\equiv\{w_{ijc}\}$ 
    \STATE {\bfseries Input:} side info $z$, base predictions $p \in[\eps;1-\eps]^{K_0-1}$
    \STATE {\bfseries Input:} binary target $x$, learning rate $\eta \in (0,1)$
    \STATE {\bfseries Input:} boolean $update$ (controls if we learn or not)
    \STATE {\bfseries Output:} estimate of $\mathbb{P}[x=1 \;|\; z, p]$
    \vspace{0.4em}
    \STATE  $p_0 \leftarrow (\beta, p_{1}, p_{2}, \dots, p_{K_0-1})$
    \vspace{0.2em}
    \FOR[loops over layers]{$i \in \{ 1,\dots,L \}$}
        \STATE  $p_{i0} \leftarrow \beta$
        \FOR[loops over neurons]{$j \in \{ 1, \dots, K_i \}$}
            \STATE $p_{ij} \leftarrow \textsc{clip}_{\eps}^{1-\eps} \left[ \sigma \left( w_{ij c_{ij}(z)} \cdot \sigma^{-1}(p_{i-1}) \right) \right]$
            \IF{$\text{update}$}
                \STATE $\Delta_{ij} \leftarrow - \eta \left( p_{ij} - x \right) \sigma^{-1} (p_{i-1})$
                \STATE $w_{ij c_{ij}(z)} \leftarrow \textsc{clip}_{-b}^b[w_{ij c_{ij}(z)} + \Delta_{ij}]$
            \ENDIF
        \ENDFOR
    \ENDFOR
    \vspace{0.2em}
    \RETURN $p_{L1}$
\end{algorithmic}
\end{algorithm}

A single prediction step, as well as a single step of learning using Online Gradient Descent, can be implemented via a single forward pass of the network as shown in Algorithm \ref{algo:updategln}.
Here we make use of a subroutine $\text{{\sc clip}}_\eps^{1-\eps}[x] := \min \left\{ \max(x, \eps), 1-\eps \right\}$.
Generating a prediction requires computing the active contexts from the given side information for each neuron, and then performing $L$ matrix-vector products.
Under the assumption that multiplying a $m \times n$ by $n \times 1$ pair of matrices takes $O(mn)$ work, the total time complexity to generate a single prediction is $O ( \sum_{i=1}^L K_i K_{i-1} )$ for the matrix-vector products, which in typical cases will dominate the overall runtime.
Note that updating the weights does not affect this complexity.

\begin{figure*}[t!]
	\begin{center}
		\includegraphics[width=0.9 \linewidth]{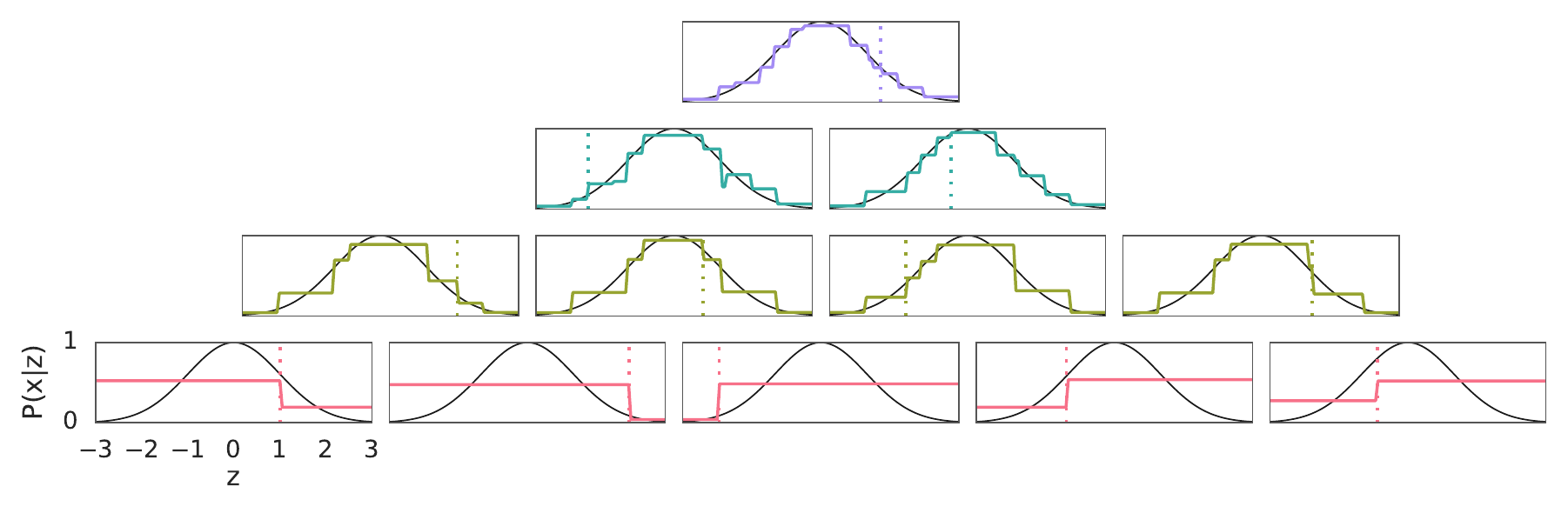}
	\end{center}
	\vspace{-2em}
	\caption{Output of a four layer network with random half-space contexts after training to convergence. Each box represents a non-bias neuron in the network, the function to fit is shown in black, and the output distribution learnt by each neuron 
		is shown in colour (for example, red for the first layer and purple for the top-most neuron). All axes are identical, as labeled in the bottom left neuron. The dashed coloured lines 
		represent the sampled hyperplane for each neuron.}
	\label{fig:gln_at_convergence}
	\vspace{-1em}
\end{figure*}

\paradot{Random halfspace sampling}
\label{sec:halfspace_sampling}
Here we describe how we generate a diverse set of halfspace context functions in practice.
As we are interested in higher dimensional applications, it is necessary to sample hyperplanes in a manner that addresses the curse of dimensionality. Consider a halfspace context function: $c(z ; v,b) = 1$ if $z\cdot v \geq b$; or $0$ otherwise. To sample $v$, we first generate an i.i.d. random vector $x = (x_1, ..., x_d)$ of dimension $d$, with each component of $x$ distributed according to the unit normal $\mathcal{N}(0,1)$, and then divide by its $2$-norm, giving us a vector $v = x / ||x||_2$. This scheme uniformly samples points from the surface of a unit sphere. 
The scalar $b$ is sampled directly from a standard normal distribution.

The motivation for this approach is two-fold. First, With large $d$, the hyperplanes defining each half-space are orthogonal with high probability; i.e. this choice should help to chop the data up in complementary ways given a limited number of gates.
Second, suppose we have a set of $m$ different gating functions $c_i(z ; v_i, b_i)$ for $1$ to $m$. 
Now consider the binary vector: 
    $g = ( c_1(z ; v_1, b_1), ..., c_m(z ; v_m, b_m) )$.
This \emph{signature} vector $g$ of input $z$ has the property \citep{simhash} that different $z$'s which are close in terms of cosine similarity will map to similar signatures. For a GLN, this gives rise to the desirable property that inputs close in cosine distance will map to similar products of data dependent matrices, i.e. they will predict similarly. 

\paradot{On convergence properties and rates for GLNs}
\label{sec:GLNrates}
Asymptotic convergence results for GLNs on i.i.d.\ data can be proven.
On-average within each context cell, 
the prediction converges to the true expected output/probability,
and a sufficiently large GLN can represent the true target probabilities arbitrarily well \cite{Veness:17}.
For example, Figure \ref{fig:gln_at_convergence} shows the converged predictions when using a small GLN to fit a simple parametrized density function. Of course while asymptotic convergence is a useful sanity check for any model, 
it tells us little about practical finite-time performance.
Below we will outline how (good) convergence rates may be obtained for fixed finite sized GLNs to the best locally learnable approximation. Precisely obtaining such bounds is outside the scope of this paper.

For a single gated neuron, one can show \citep{Veness:17} 
that Online Gradient Descent (OGD) \citep{zinkevich03}
with a learning rate proportional to $1/\sqrt{t}$ has total regret of $O(\sqrt{T})$ 
with respect to the best $w^*\in\cal W$ chosen in hindsight.
The loss function $\ell_t$ is exp-concave, 
so Online Newton Step \cite{Hazan2007LogarithmicRA} can improve the regret to $O(\log{T})$, 
but is computationally more expensive.
If the expected loss $\ell(w):=\mathbb{E}[\ell_t(w)]$ were strongly convex, 
then Stochastic Gradient Descent (SGD) with i.i.d.\ sampling
and a learning rate proportional to $1/t$ would also achieve a regret of $O(\log{T})$,
Unfortunately $\ell_t$ is flat in all directions orthogonal to $\logit(p_t)$,
hence not strongly convex.
But since $\ell_t$ is exp-concave (strongly convex in gradient direction), 
this makes $\ell(w)$ strongly convex in the linear subspace of ${\cal W}\subset\mathbb{R}^d$
spanned by ${\cal S}:=\text{Span}(\logit(p_1),...,\logit(p_n))$. 
For sample size $n$ larger than $d$ it is plausible that ${\cal S}=\mathbb{R}^d$.
Even if not, all $\ell_t$ are exactly constant in directions orthogonal to $\cal S$,
hence the gradient lies in $\cal S$. 
Since $\ell$ is strongly convex in $\cal S$,
SGD will still achieve a regret of $O(\log T)$.

The above implies that the time-averaged weights $\bar w_T$ 
have an instantaneous regret of $O(\log T~/T)$,
and even $O(1/T)$ can be achieved \citep[Thm.6.2]{Bubeck:15}.
In general, OGD algorithms can be converted to achieve these rates even for the current weight $w_t$ \citep{Cutkosky:19},
and, indeed, SGD achieves the former even unmodified \citep{Shamir:13}.
By strong convexity on $\cal S$, 
this implies that the (time-averaged) weights (or at least the outputs) converge with a rate of $\tilde O(t^{-1/2})$.

Hence after time $\tilde O(1/\eps^2)$ the output of the first layer has converged within $O(\eps)$,
after which the input to the next layer becomes approximately i.i.d.
A similar analysis should then be possible for the second layer,
and so on. With an appropriately delayed learning rate decay, 
this should lead to an overall time bound of $O(L/\eps^2)$ 
to achieve $\eps$-approximation.
For these reasons we use gradient descent with a learning rate proportional to $1/t$ in our experiments described in Section~\ref{sec:obench}.

\section{Empirical Capacity of GLNs}\label{sec:capacity}

Contemporary neural networks have the desirable capability to approximate arbitrary continuous functions given almost any reasonable activation function  \citep[and others]{Hor91}. GLNs share this property so long as the context capacity is sufficiently expressive. Moreover,~\cite{Veness:17} prove that this capacity is \textit{effective} in the sense that gradient descent will eventually find the best feasible approximation. This property is not shared by neural networks trained by backpropagation; it is possible to demonstrate the existence of such weights, but not to guarantee that gradient descent (or any other practical algorithm) will find them.
Here we demonstrate the capacity of GLNs in practice by measuring their ability to fit random labelled data.

\begin{figure}[t!]
	\centering
	\includegraphics[width=0.9 \linewidth]{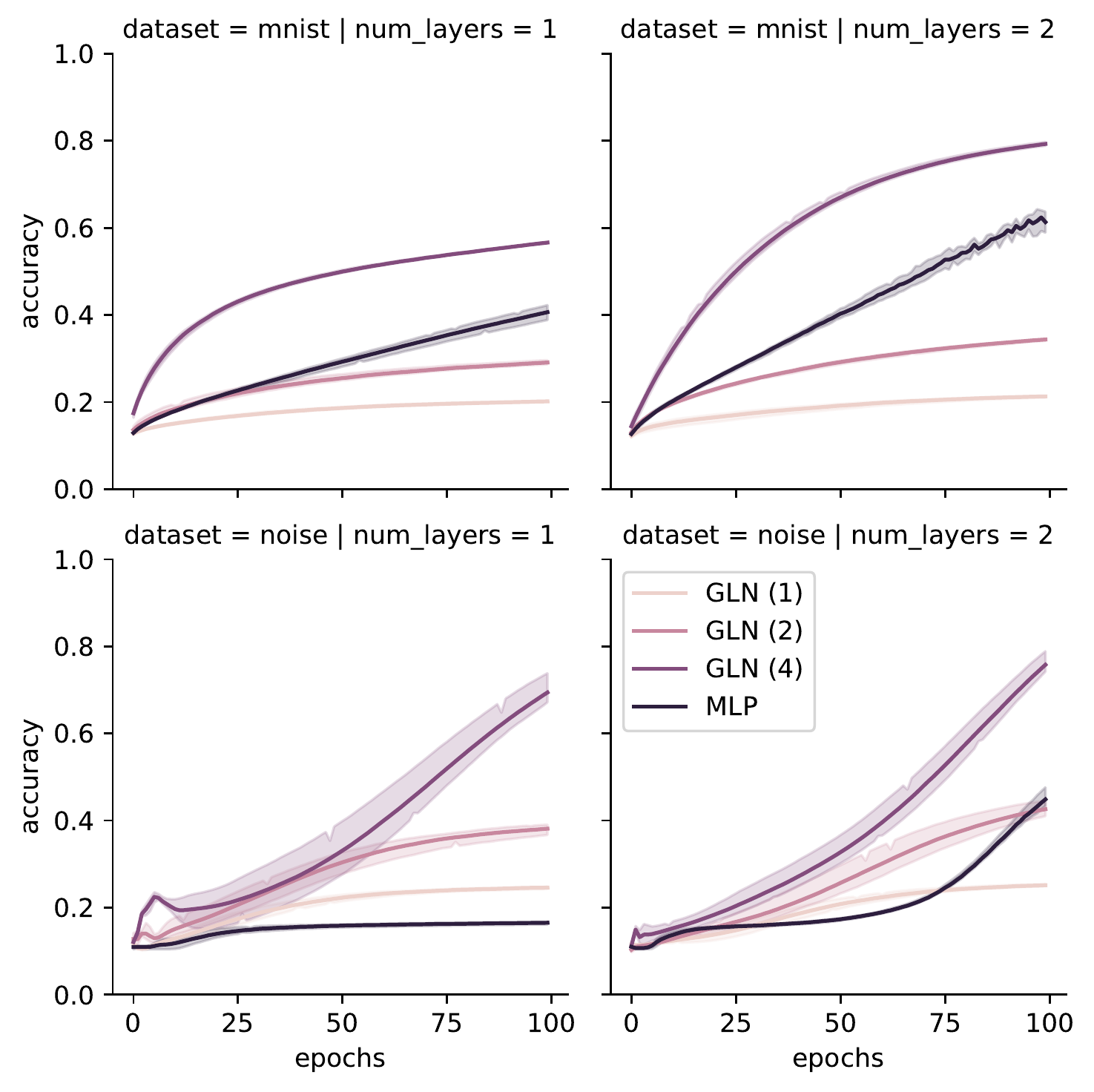}
	\vspace{-1em}
	\caption{Empirical capacity of GLNs (context dimension showed in parentheses) versus an MLP baseline. Top row represents the MNIST dataset with shuffled labels. Bottom row represents a dataset of uniform noise of the same size and shape.
	\label{fig:capacity_results}}
 	\vspace{-2em}
\end{figure}

 We ran two sets of experiments: first, using the standard MNIST dataset with shuffled labels; and second, replacing the MNIST images with uniform noise of the same shape and dataset length. These results are presented in Figure~\ref{fig:capacity_results} compared to an MLP baseline in an equivalent one-vs-all configuration. For GLNs, we select a fixed layer width of 128 and vary both the context dimension and number of layers. For the MLP, we select the number of \emph{neurons} such that the total number of \emph{weights} in the network is equivalent to a GLN with context dimension 4 (the largest considered). The GLN was trained with learning rate $10^{-4}$ and the MLP using the Adam optimizer~\cite{kingma2014adam} with learning rate $10^{-5}$, both selected by conducting a sweep over learning rates from $10^{-1}$ to $10^{-6}$. It is evident from Figure~\ref{fig:capacity_results} that GLNs have comparable capacity to an equivalently sized MLP in practice, with their ability to memorize training data scaling in both the number of neurons and context dimension.

\section{Linear Interpretability of GLNs}

\begin{figure}[t!]
\begin{minipage}[b]{0.7\columnwidth}
  \includegraphics[scale=0.2]{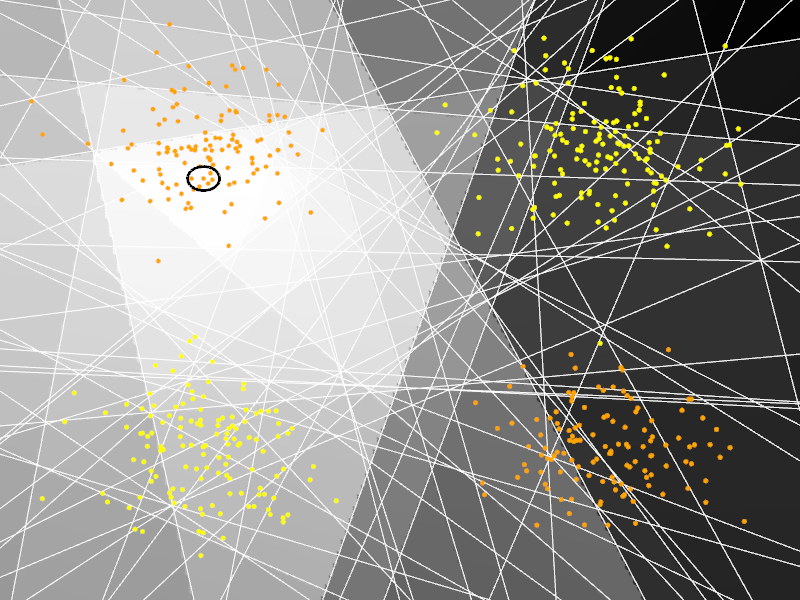}
\end{minipage}%
\hfill%
\begin{minipage}[b]{0.28\columnwidth}
	\caption{The effect of a single noisy XOR update (circled) on the decision boundaries of a halfspace gated GLN. Sampled hyperplanes for each gate are shown in white.}\label{fig:bias}
\end{minipage}
\end{figure}

In the case where we have an $L$ layer GLN, with $K_i$ neurons on layer $i$, and with input $p_0$ of dimension $K_0$ and side information $z$, the RHS of Equation \ref{eq:linear} can be written as
\begin{equation*}
\sigma\Bigl( \underbrace{ W_{L}(z) \, W_{L-1}(z) \, \dots \, W_{1}(z)}_{\text{multilinear polynomial of degree L} }\logit(p_0) \Bigr),
\end{equation*}
where each matrix $W_i(z)$ is of dimension $K_i \times K_{i-1}$, with the $j$th row constituting the active weights (as determined by the gating) for the $j$th neuron in layer $i$.
This formulation is convenient for many reasons. First, it allows us to reason about the inductive bias of GLNs by observing that the product of matrices collapses to a multilinear polynomial in the learnt weights, i.e. the depth and shape of the network directly influences how a GLN will generalize. A visual example of the change in decision boundaries resulting from a single halfspace gated GLN update is shown in Figure \ref{fig:bias} for the noisy XOR problem. The magnitude of the change is largest within the convex polytope containing the training point, and decays with respect to the remaining convex polytopes according to how many halfspaces they share with the containing convex polytope. This makes intuitive sense, as since the weight update is local, each row of $W_i(z)$ is pushed in the direction to better explain the data independently of each other. Therefore one should think of a halfspace gated GLN as a smoothing technique -- input points which cause similar gating activation patterns must have similar outputs.

\begin{figure}[h!]
	\centering
	\includegraphics[width=\linewidth]{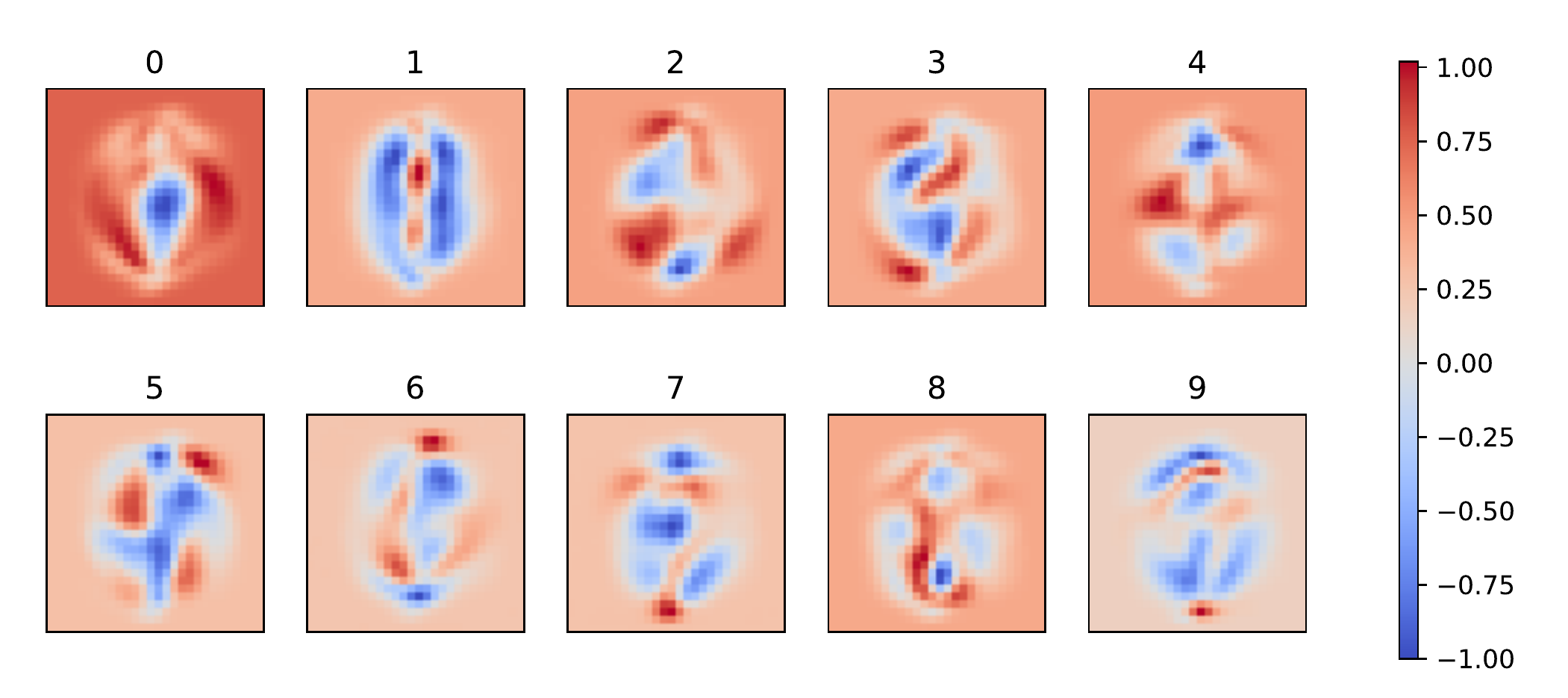}
	\vspace{-5ex}
	\caption{Saliency maps for constituent GLN binary classifiers of one-vs-all MNIST classifier after a single training epoch.\label{fig:gln_saliency}}
	\vspace{-2em}
\end{figure}

\begin{figure*}[t!]
	\begin{center}
		\includegraphics[width= \linewidth]{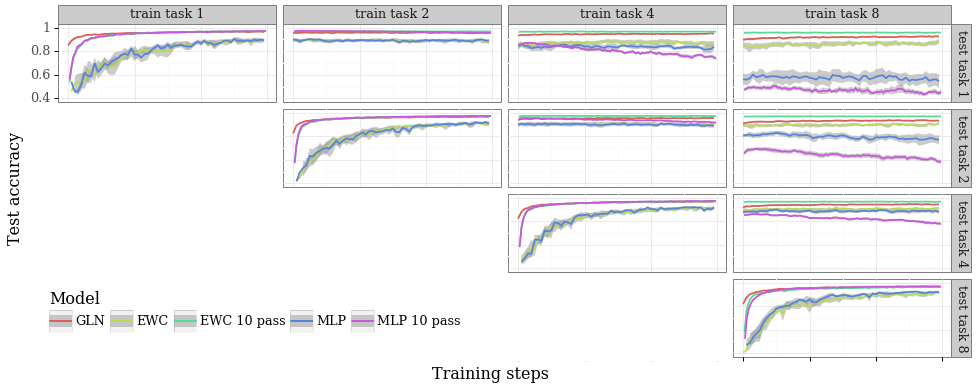}
	\end{center}
	\vspace{-2em}
	\caption{Retention results for permuted MNIST. Models are trained sequentially on 8 tasks (rows) and evaluated on all previously encountered tasks (columns). For example, the top-right plot indicates performance on Task 1 after being trained sequentially on Tasks 1 to 8 inclusive (not all tasks shown). Each model only trains for one epoch per task, with the exception of ``EWC 10 pass'' and ``MLP 10 pass'' (shrunken 10-fold on x axis). Error bars denote 95\% confidence levels over 10 random seeds.}
	\label{fig:forgetting}
	\vspace{-1em}
\end{figure*}

Aside from reasoning about the inductive bias, the above formulation provides a convenient mechanism for interpreting the learnt weights of a trained GLN. Contemporary neural networks have been criticized by some as ``black boxes'' that are notoriously difficult to interpret~\cite{yosinski2015understanding, zhang2018visual}. Despite their high discrimination power, this can prove problematic for learning and debugging efficiently at the semantic level as well as for deployment in safety-critical real-world applications. This has led to the development of gradient-based methods for post-hoc network analysis~\cite{simonyan2013deep}. Such methods are not necessary for GLNs; for a given input, the collapsed multilinear polynomial of degree L is a weight vector of the same dimension (since $W_L(z)$ has 1 row and $W_1(z)$ has $K_0$ columns) as the inputs and provides a natural formulation for intuitive saliency maps without any further computational expense. An example of the obtained saliency maps are provided in Figure~\ref{fig:gln_saliency} for a one-versus-all GLN trained as an MNIST classifier. One can clearly see that the characteristic shape of each hand-written character is preserved.

\section{Resilience to Catastrophic Forgetting}

Humans are able to acquire new skills throughout life seemingly without compromising their ability to solve previously learnt tasks. Contemporary neural networks do not share this ability; if a network is trained on a task $A$ and these weights are used to initialize training for a new task $B$, the ability to solve $A$ rapidly degrades as training progresses on $B$. This phenomenon of ``catastrophic forgetting" has been well studied for decades~\cite{carpenter1988, McCloskey1989, Robins1995} but continues to limit the applicability of neural networks in continual or lifelong learning scenarios.

Similar to the problem of model interpretability, many algorithms have been developed that augment standard training by backpropagation to address catastrophic forgetting. These methods typically fall into two main categories. The first approach involves replaying previously seen tasks during training using one of many heuristics~\cite{Robins1995, Caruana1997, Rebuffi2017}. The other common category involves explicitly maintaining additional sets of model parameters related to previously learnt tasks. Examples include freezing a subset of weights~\cite{Donahue2013, Razavian2014}, dynamically adjusting learning rates~\cite{Girshick2014} or augmenting the loss with regularization terms with respect to past parameters~\cite{Kirkpatrick17, Friedemann17, Schwarz2018}. A limitation of these approaches (aside from additional algorithmic and computational complexity) is that they require task boundaries to be provided or accurately inferred.

Unlike contemporary neural networks, we demonstrate that the halfspace-gated GLN architecture and learning rule is naturally robust to catastrophic forgetting without any modifications or knowledge of task boundaries. We focus on the pixel-permuted MNIST continual learning benchmark of \cite{Goodfellow2013, Kirkpatrick17}, which involves training on a sequence of different tasks where each task is obtained from a different random permutation of the input pixels. We compare the learning and retention characteristics of a GLN against an MLP baseline (of equal number of neurons, using dropout as per the original paper) with and without elastic weight consolidation (EWC)~\cite{Kirkpatrick17}, which is a highly-effective method explicitly designed to prevent catastrophic forgetting by storing parameters of previously seen tasks. 

Our results are presented in Figure~\ref{fig:forgetting}. As we train our models on a growing number of sequential tasks (rows), the performance on all previously learnt tasks (columns) is evaluated. Note that the plotted task indices are not contiguous. It is evident that the GLN outperforms EWC in terms of both initial single-task learning (diagonal) and retention when both are trained for a single pass. Only when EWC is trained for multiple (ten) passes over the data does it exhibit superior performance to a vanilla GLN. In all tests, the GLN substantially outperforms the standard MLP without EWC.

To gain some intuition as to why GLNs are resilient to catastrophic interference, recall from Section \ref{sec:halfspace_sampling} that inputs close in terms of cosine similarity will give rise to similar data dependent weight matrices.
Since each task-specific cluster of examples is far from each other in signature space, the amount of interference between tasks is significantly reduced, with the gating essentially acting as an implicit weight hashing mechanism.

\section{Online Benchmarking}\label{sec:obench}

\paragraph{MNIST Classification.}

First we explore the use of GLNs for online (single-pass) classification of the deskewed MNIST dataset~ \cite{Lecun98,deskew}. We use 10 GLNs to construct a one-vs-all classifier, each consisting of 128 neurons per layer with context dimension 4. The learning rate at each step $t$ was set to $\min\{100/t, 0.01\}$. We find that the GLN is capable of impressive online performance, achieving $98\%$ accuracy in a single pass of the training data. 

\paragraph{UCI Dataset Classification.}

\begin{figure}[t!]
	\centering
	\includegraphics[width=0.9 \linewidth]{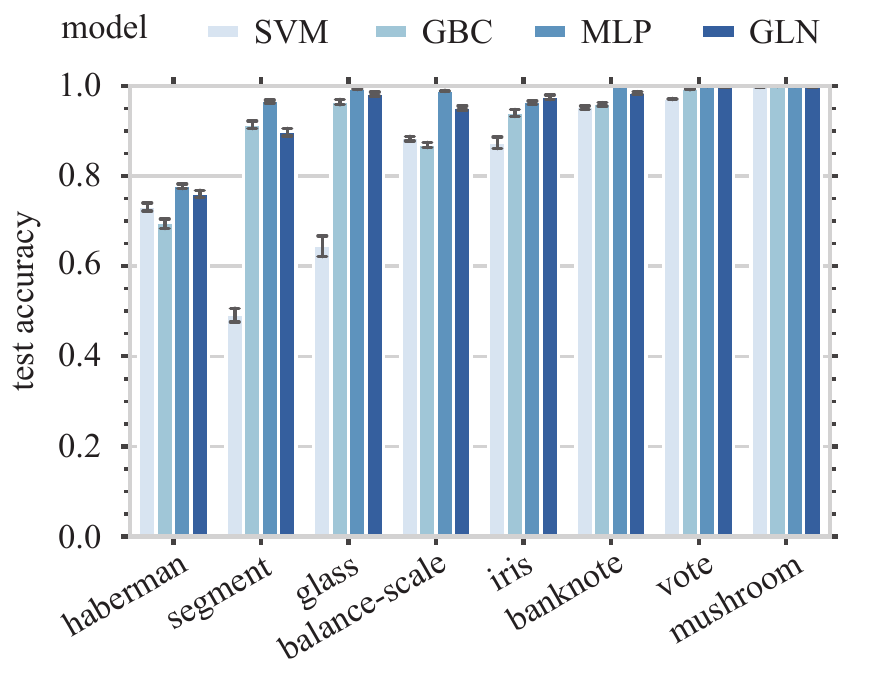}
	\vspace{-0.5em}
	\caption{Online (single-pass) GLN classification accuracy on a selection UCI datasets, compared to three contemporary batch methods (Support Vector Machine, Gradient Boosting for Classification, Multi-Layer Perceptron) trained for 100 epochs.\label{fig:gln_results}}
	\vspace{-1em}
\end{figure}

We next compare GLNs to a variety of general purpose batch learning techniques (SVMs, Gradient Boosting for Classification, MLPs) in small data regimes on a selection of standard UCI datasets.
A 1000-500 neuron GLN with context-dimension 8 was trained with a \emph{single pass} over 80\% of instances and evaluated with frozen weights on the remainder. 
The comparison MLP used ReLU activations and the same number of weights, and was trained for \emph{100 epochs} using the Adam optimizer~\cite{kingma2014adam} with learning rate $0.001$ and batch size 32.
The SVM classifier used a radial basis function kernel $K(x, x')=\exp \{-\gamma \norm{x-x'}^2 \}$ with $\gamma=1/d$, where $d$ is the input dimension.
The GBC classifier was an ensemble of 100 trees of maximum depth 3 with a learning rate of $0.1$.
The mean and stderr over 100 random train/test splits are shown in the leftmost graph of Figure \ref{fig:gln_results}.
Here we see that the single-pass GLN is competitive with the best of the batch learning results on each domain.

\paragraph{MNIST Density Modelling.}
Our final result is to use GLNs and image specific gating to construct an online image density model for the binarized MNIST dataset \citep{larochelle11a}, a standard benchmark for image density modeling.
By exploiting the chain rule $\mathbb{P}(X_{1:d})=\prod_{i=1}^{d} \mathbb{P}(X_i \, | \, X_{<i})$ of probability, we constructed an autoregressive density model over the $28\times28$ dimensional binary space by using 784 GLNs to model the conditional distribution for each pixel; a row-major ordering was used to linearize the two dimensional pixel locations. Running our method online (i.e. a single pass of the concatenated training, validation and test sets) gave an average loss of $79.0$ nats per image across the test data, and $80.74$ nats per image if we held the parameters fixed upon reaching the test set. These results are close to state of the art \citep{VanDenOord2016} of any batch trained density model which outputs exact probabilities.

From an MDL or compression perspective, our density modelling results are significantly stronger in the sense that we could couple our model to an adaptive arithmetic decoder and reproduce the original data from a file much smaller than the original input. Contemporary batch trained density models do not have this property; to make a fair comparison, they would need to first encode the parameters of the model before encoding the subsequent compressed data, and state-of-the-art batch train density models typically have compressed size many orders of magnitude larger than the original data.

\section{Conclusion}

We have introduced a new family of general purpose neural architectures, Gated Linear Networks, and studied the desirable characteristics that follow from their use of data-dependent gating and local credit assignment. 
Their fast online learning properties, easy interpretability, and excellent robustness to catastrophic forgetting in continual learning settings makes them an interesting and complementary alternative to contemporary deep learning approaches.

\bibliographystyle{alpha}

\begin{thebibliography}{VDO} 

\bibitem[B{\etalchar{+}}15]{Bubeck:15}
S{\'e}bastien Bubeck et~al.
\newblock Convex optimization: Algorithms and complexity.
\newblock {\em Foundations and Trends{\textregistered} in Machine Learning},
  8(3-4):231--357, 2015.

\bibitem[BFH{\etalchar{+}}18]{jax2018github}
James Bradbury, Roy Frostig, Peter Hawkins, Matthew~James Johnson, Chris Leary,
  Dougal Maclaurin, and Skye Wanderman-Milne.
\newblock {JAX}: composable transformations of {P}ython+{N}um{P}y programs,
  2018.
  
\bibitem[BH89]{Baldi1989}
P.~Baldi and K.~Hornik.
\newblock Neural networks and principal component analysis: Learning from
  examples without local minima.
\newblock {\em Neural Networks}, 2(1):53--58, January 1989.

  
\bibitem[BHQK20]{rlax2020github}
David Budden, Matteo Hessel, John Quan, and Steven Kapturowski.
\newblock {RL}ax: {R}einforcement {L}earning in {JAX}, 2020.

\bibitem[Car97]{Caruana1997}
Rich Caruana.
\newblock Multitask learning.
\newblock {\em Machine Learning}, 28(1):41--75, Jul 1997.

\bibitem[CG88]{carpenter1988}
G.~A. {Carpenter} and S.~{Grossberg}.
\newblock The art of adaptive pattern recognition by a self-organizing neural
  network.
\newblock {\em Computer}, 21(3):77--88, March 1988.

\bibitem[Cha02]{simhash}
M.S. Charikar.
\newblock Similarity estimation techniques from rounding algorithms.
\newblock {\em Conference Proceedings of the Annual ACM Symposium on Theory of
  Computing}, pages 380--388, 01 2002.

\bibitem[Cut19]{Cutkosky:19}
Ashok Cutkosky.
\newblock Anytime {{Online}}-to-{{Batch}}, optimism and acceleration.
\newblock In Kamalika Chaudhuri and Ruslan Salakhutdinov, editors, {\em
  Proceedings of the 36th International Conference on Machine Learning},
  volume~97 of {\em Proceedings of Machine Learning Research}, pages
  1446--1454, {Long Beach, California, USA}, June 2019. {PMLR}.

\bibitem[DJV{\etalchar{+}}13]{Donahue2013}
Jeff Donahue, Yangqing Jia, Oriol Vinyals, Judy Hoffman, Ning Zhang, Eric
  Tzeng, and Trevor Darrell.
\newblock Decaf: {A} deep convolutional activation feature for generic visual
  recognition.
\newblock {\em CoRR}, abs/1310.1531, 2013.

\bibitem[GDDM14]{Girshick2014}
Ross Girshick, Jeff Donahue, Trevor Darrell, and Jitendra Malik.
\newblock Rich feature hierarchies for accurate object detection and semantic
  segmentation.
\newblock {\em 2014 IEEE Conference on Computer Vision and Pattern
  Recognition}, Jun 2014.

\bibitem[GMX{\etalchar{+}}13]{Goodfellow2013}
Ian~J. Goodfellow, Mehdi Mirza, Da~Xiao, Aaron Courville, and Yoshua Bengio.
\newblock An empirical investigation of catastrophic forgetting in
  gradient-based neural networks, 2013.

\bibitem[GW17]{deskew}
Dibya Ghosh and Alvin Wan, 2017.\\
\newblock https://fsix.github.io/mnist/

\bibitem[HAK07]{Hazan2007LogarithmicRA}
Elad Hazan, Amit Agarwal, and Satyen Kale.
\newblock Logarithmic regret algorithms for online convex optimization.
\newblock {\em Machine Learning}, 69:169--192, 2007.

\bibitem[Haz16]{Hazan:16}
Elad Hazan.
\newblock Introduction to online convex optimization.
\newblock {\em Foundations and Trends in Optimization}, 2(3-4):157--325, 2016.

\bibitem[HCNB20]{haiku2020github}
Tom Hennigan, Trevor Cai, Tamara Norman, and Igor Babuschkin.
\newblock {H}aiku: {S}onnet for {JAX}, 2020.

\bibitem[Hin02]{Hinton2002}
Geoffrey~E. Hinton.
\newblock Training products of experts by minimizing contrastive divergence.
\newblock {\em Neural Computation}, 14(8):1771--1800, August 2002.

\bibitem[Hor91]{Hor91}
Kurt Hornik.
\newblock Approximation capabilities of multilayer feedforward networks.
\newblock {\em Neural networks}, 4(2):251--257, 1991.

\bibitem[KB14]{kingma2014adam}
Diederik~P Kingma and Jimmy Ba.
\newblock Adam: A method for stochastic optimization.
\newblock {\em arXiv preprint arXiv:1412.6980}, 2014.

\bibitem[Kno17]{cmix}
Byron Knoll, 2017.\\
\newblock http://www.byronknoll.com/cmix.html

\bibitem[KPR{\etalchar{+}}17]{Kirkpatrick17}
James Kirkpatrick, Razvan Pascanu, Neil Rabinowitz, Joel Veness, Guillaume
  Desjardins, Andrei~A. Rusu, Kieran Milan, John Quan, Tiago Ramalho, Agnieszka
  Grabska-Barwinska, Demis Hassabis, Claudia Clopath, Dharshan Kumaran, and
  Raia Hadsell.
\newblock Overcoming catastrophic forgetting in neural networks.
\newblock {\em Proceedings of the National Academy of Sciences},
  114(13):3521--3526, 2017.

\bibitem[LBBH98]{Lecun98}
Yann Lecun, Léon Bottou, Yoshua Bengio, and Patrick Haffner.
\newblock Gradient-based learning applied to document recognition.
\newblock In {\em Proceedings of the IEEE}, pages 2278--2324, 1998.

\bibitem[LM11]{larochelle11a}
Hugo Larochelle and Iain Murray.
\newblock The neural autoregressive distribution estimator.
\newblock In Geoffrey Gordon, David Dunson, and Miroslav Dudík, editors, {\em
  Proceedings of the Fourteenth International Conference on Artificial
  Intelligence and Statistics}, volume~15 of {\em Proceedings of Machine
  Learning Research}, pages 29--37, Fort Lauderdale, FL, USA, 11--13 Apr 2011.
  PMLR.

\bibitem[Mah00]{Mahoney2000}
Matthew Mahoney.
\newblock Fast text compression with neural networks.
\newblock {\em AAAI}, 2000.

\bibitem[Mah05]{Mahoney2005}
Matthew Mahoney.
\newblock Adaptive weighing of context models for lossless data compression.
\newblock {\em Technical Report, Florida Institute of Technology CS}, 2005.

\bibitem[Mah13]{Mahoney2013}
Matthew Mahoney.
\newblock {\em Data Compression Explained}.
\newblock Dell, Inc, 2013.

\bibitem[Mat12]{Mattern12}
Christopher Mattern.
\newblock Mixing strategies in data compression.
\newblock In {\em 2012 Data Compression Conference, Snowbird, UT, USA, April
  10-12}, pages 337--346, 2012.

\bibitem[Mat13]{Mattern13}
Christopher Mattern.
\newblock Linear and geometric mixtures - analysis.
\newblock In {\em 2013 Data Compression Conference, {DCC} 2013, Snowbird, UT,
  USA, March 20-22, 2013}, pages 301--310, 2013.

\bibitem[MC89]{McCloskey1989}
Michael McCloskey and Neal~J. Cohen.
\newblock Catastrophic interference in connectionist networks: The sequential
  learning problem.
\newblock volume~24 of {\em Psychology of Learning and Motivation}, pages 109
  -- 165. Academic Press, 1989.

\bibitem[MP69]{minsky69perceptrons}
Marvin Minsky and Seymour Papert.
\newblock {\em Perceptrons: An Introduction to Computational Geometry}.
\newblock MIT Press, Cambridge, MA, USA, 1969.

\bibitem[OWR{\etalchar{+}}19]{peedrrrrooo}
Pedro~A. Ortega, Jane~X. Wang, Mark Rowland, Tim Genewein, Zeb Kurth{-}Nelson,
  Razvan Pascanu, Nicolas Heess, Joel Veness, Alexander Pritzel, Pablo
  Sprechmann, Siddhant~M. Jayakumar, Tom McGrath, Kevin Miller,
  Mohammad~Gheshlaghi Azar, Ian Osband, Neil~C. Rabinowitz, Andr{\'{a}}s
  Gy{\"{o}}rgy, Silvia Chiappa, Simon Osindero, Yee~Whye Teh, Hado van Hasselt,
  Nando de~Freitas, Matthew Botvinick, and Shane Legg.
\newblock Meta-learning of sequential strategies.
\newblock {\em CoRR}, abs/1905.03030, 2019.

\bibitem[RASC14]{Razavian2014}
Ali~Sharif Razavian, Hossein Azizpour, Josephine Sullivan, and Stefan Carlsson.
\newblock Cnn features off-the-shelf: An astounding baseline for recognition.
\newblock {\em 2014 IEEE Conference on Computer Vision and Pattern Recognition
  Workshops}, Jun 2014.

\bibitem[RKSL17]{Rebuffi2017}
Sylvestre-Alvise Rebuffi, Alexander Kolesnikov, Georg Sperl, and Christoph~H.
  Lampert.
\newblock icarl: Incremental classifier and representation learning.
\newblock {\em 2017 IEEE Conference on Computer Vision and Pattern Recognition
  (CVPR)}, Jul 2017.

\bibitem[Rob95]{Robins1995}
Anthony~V. Robins.
\newblock Catastrophic forgetting, rehearsal and pseudorehearsal.
\newblock {\em Connect. Sci.}, 7:123--146, 1995.

\bibitem[SLC{\etalchar{+}}18]{Schwarz2018}
Jonathan Schwarz, Jelena Luketina, Wojciech~M. Czarnecki, Agnieszka
  Grabska-Barwinska, Yee~Whye Teh, Razvan Pascanu, and Raia Hadsell.
\newblock Progress \& compress: A scalable framework for continual learning,
  2018.

\bibitem[SMG13]{SaxeMG13}
Andrew~M. Saxe, James~L. McClelland, and Surya Ganguli.
\newblock Exact solutions to the nonlinear dynamics of learning in deep linear
  neural networks.
\newblock {\em CoRR}, abs/1312.6120, 2013.

\bibitem[SVZ13]{simonyan2013deep}
Karen Simonyan, Andrea Vedaldi, and Andrew Zisserman.
\newblock Deep inside convolutional networks: Visualising image classification
  models and saliency maps.
\newblock {\em arXiv preprint arXiv:1312.6034}, 2013.

\bibitem[SZ13]{Shamir:13}
Ohad Shamir and Tong Zhang.
\newblock Stochastic gradient descent for non-smooth optimization: Convergence
  results and optimal averaging schemes.
\newblock In {\em International conference on machine learning}, pages 71--79,
  2013.

\bibitem[VDOKK16]{VanDenOord2016}
A\"{a}ron Van Den~Oord, Nal Kalchbrenner, and Koray Kavukcuoglu.
\newblock Pixel recurrent neural networks.
\newblock In {\em Proceedings of the 33rd International Conference on
  International Conference on Machine Learning - Volume 48}, ICML'16, pages
  1747--1756. JMLR.org, 2016.

\bibitem[VLB{\etalchar{+}}17]{Veness:17}
Joel Veness, Tor Lattimore, Avishkar Bhoopchand, Agnieszka Grabska-Barwinska,
  Christopher Mattern, and Peter Toth.
\newblock Online learning with gated linear networks.
\newblock {\em arXiv preprint arXiv:1712.01897}, 2017.

\bibitem[YCN{\etalchar{+}}15]{yosinski2015understanding}
Jason Yosinski, Jeff Clune, Anh Nguyen, Thomas Fuchs, and Hod Lipson.
\newblock Understanding neural networks through deep visualization.
\newblock {\em arXiv preprint arXiv:1506.06579}, 2015.

\bibitem[Zin03]{zinkevich03}
Martin Zinkevich.
\newblock Online convex programming and generalized infinitesimal gradient
  ascent.
\newblock In {\em Machine Learning, Proceedings of the Twentieth International
  Conference {(ICML} 2003), August 21-24, 2003, Washington, DC, {USA}}, pages
  928--936, 2003.

\bibitem[ZPG17]{Friedemann17}
Friedemann Zenke, Ben Poole, and Surya Ganguli.
\newblock Continual learning through synaptic intelligence.
\newblock In {\em Proceedings of the 34th International Conference on Machine
  Learning - Volume 70}, ICML’17, page 3987–3995. JMLR.org, 2017.

\bibitem[ZZ18]{zhang2018visual}
Quan-shi Zhang and Song-Chun Zhu.
\newblock Visual interpretability for deep learning: a survey.
\newblock {\em Frontiers of Information Technology \& Electronic Engineering},
  19(1):27--39, 2018.






\end{thebibliography}
\newcommand{\etalchar}[1]{$^{#1}$}

\clearpage
\appendix

\section*{Additional Experiment Details}

\subsection*{Permuted-MNIST}

This section describes the implementation details and hyperparameters for each experiment.

\textbf{GLN.~} We use one-vs-all GLNs composed of $[100, 25, 1]$ neurons per layer and a context dimension of $6$. The output of the network is determined by the last neuron.

\textbf{MLP and EWC.~} We use a ReLU network with $[1000, 250, 10]$ neurons per layer. We use Adam \cite{kingma2014adam} to optimize the cross-entropy loss using mini-batches of $20$ data points. For EWC, we draw $100$ samples for computing the Fisher matrix diagonals. We also optimize the constant that trades-off remembering and learning via grid-search, and denote it with $\lambda$ in Table~\ref{table:forgetting_params}.

\begin{table}[h]
\begin{center}
\begin{tabular}{ c || c | c | c } 
\textbf{Model} & log-learning rate & dropout & $\log \lambda$ \\
\hline
GLN & -4, -3, \textbf{-2}, -1  & --  & --\\
MLP & -6, -5, \textbf{-4}, -3  & Yes, \textbf{No} & --\\
EWC & -6, -5, \textbf{-4}, -3  & Yes, \textbf{No} & 2, \textbf{3}, 4\\
\end{tabular}
\end{center}
\caption{Parameters explored during grid search. The best parameters (shown in bold) are the ones that maximize the average accuracy over 10 random seeds. Logarithms are in base 10.}
\label{table:forgetting_params}
\end{table}

\subsection*{MNIST saliency map experimental details}
We use one-vs-all GLNs composed of $[50, 25, 1]$ neurons per layer and a context dimension of $4$. The output of the network is determined by the last neuron. Constant learning rate of $10^{-2}$ is used to pass though the training data once. Learned weights for bias term are dropped from analysis. 

\subsection*{Computing Infrastructure}
All experiments were ran on single-GPU desktop PCs. Models are implemented in JAX \cite{jax2018github} making use of Haiku \cite{haiku2020github} and RLax \cite{rlax2020github}.

\end{document}